\documentclass[10pt,twocolumn,letterpaper]{article}

\usepackage{cvpr}              
\usepackage{multirow}
\usepackage{url}            
\usepackage{booktabs}       
\usepackage{amsfonts}       
\usepackage{nicefrac}       
\usepackage{microtype}      
\usepackage{xcolor}         
\usepackage{multirow}
\usepackage{graphicx}
\usepackage[accsupp]{axessibility} 
\definecolor{cvprblue}{rgb}{0.21,0.49,0.74}
\usepackage[pagebackref,breaklinks,colorlinks,allcolors=cvprblue]{hyperref}

\def\paperID{7} 
\def\confName{CVPR Workshop}
\def\confYear{2025}

\title{FM-LoRA: Factorized Low-Rank Meta-Prompting for Continual Learning}

\author{Xiaobing Yu$^{1}$, Jin Yang$^1$, Xiao Wu$^{1,2}$, Peijie Qiu$^{1,2}$, Xiaofeng Liu$^{3,4}$\\~\\
$^1$Dept. of Radiology, Washington University in St. Louis, St. Louis, USA\\
$^2$Dept. of Computer Vision, MBZUAI, Abu Dhabi, United Arab Emirates\\
$^3$Dept. of Radiology and Biomedical Imaging, Yale University, New Haven, USA\\
$^4$Dept. of Biomedical Informatics and Data Science, Yale University, New Haven, USA\\
}

\begin{document}
\maketitle
\begin{abstract}
How to continuously adapt a pre-trained model for sequential tasks with different prediction class labels and/or domains, and finally learn a generalizable model across diverse tasks is a long-lasting challenge. Continual learning (CL) has emerged as a promising approach to leverage pre-trained models (e.g., Transformers) for sequential tasks. While many existing CL methods incrementally store additional learned structures, such as Low-Rank Adaptation (LoRA) adapters or prompts—and sometimes even preserve features from previous samples to maintain performance. This leads to unsustainable parameter growth and escalating storage costs as the number of tasks increases. Moreover, current approaches often lack task similarity awareness, which further hinders the model’s ability to effectively adapt to new tasks without interfering with previously acquired knowledge. To address these challenges, we propose FM-LoRA, a novel and efficient low-rank adaptation method that integrates both a dynamic rank selector (DRS) and dynamic meta-prompting (DMP). This framework allocates model capacity more effectively across tasks by leveraging a shared low-rank subspace critical for preserving knowledge, thereby avoiding continual parameter expansion. Extensive experiments on various CL benchmarks, including ImageNet-R, CIFAR100, and CUB200 for class-incremental learning (CIL), and DomainNet for domain-incremental learning (DIL), with Transformers backbone demonstrate that FM-LoRA effectively mitigates catastrophic forgetting while delivering robust performance across a diverse range of tasks and domains.
\end{abstract}    
\section{Introduction}
\label{sec:intro}

Deep neural networks have achieved remarkable success across various areas, yet face challenges in adapting and generalizing well across a diverse array of sequential tasks with different prediction class labels and/or shifting domain environments. To cope with real-world dynamics, an intelligent system needs to incrementally acquire, update, accumulate, and exploit knowledge throughout its lifetime~\cite{wang2024comprehensive}. To address these challenges, \textbf{continual learning} (CL) has recently emerged as a promising solution. However, CL requires the model's ability to learn new tasks with different prediction labels and/or domains sequentially without forgetting previously acquired knowledge. This phenomenon, known as \textbf{catastrophic forgetting}, hampers the deployment of models in dynamic environments where tasks evolve over the model lifetime.

To address this issue, several strategies have been proposed, including rehearsal methods, weight regularization, and modular architectures. Rehearsal methods~\cite{rebuffi2017icarl,chaudhry2018efficient,lopez2017gradient} require the storage and replay of past data, leading to significant memory overhead and potential privacy issues. Weight regularization techniques~\cite{kirkpatrick2017overcoming,aljundi2018memory,zenke2017continual,li2017learning} restrict parameter updates, but often limit adaptability and flexibility in new tasks. Modular architectures~\cite{fernando2017pathnet,zhou2024expandable,yan2021dynamically} allocate dedicated modules per task, resulting in rapidly growing model complexity and reduced scalability. Furthermore, these approaches do not integrate well with large-scale pre-trained models like ViTs~\cite{dosovitskiy2020image}, as they tend to disrupt the carefully learned representations and fail to exploit the models' inherent generalization capabilities fully~\cite{wang2022dualprompt,wang2022learning}.

Recently, \textbf{parameter-efficient fine-tuning (PEFT)}~\cite{ding2023parameter} methods have emerged as scalable and efficient solutions for CL. Techniques such as Low-Rank Adaptation (LoRA)~\cite{hu2022lora} and prompt tuning~\cite{lester2021power} have gained popularity as they allow models to effectively adapt to new tasks by updating only a small subset of parameters or input embeddings. These approaches typically freeze the pre-trained model parameters and incrementally fine-tune only the newly introduced task-specific parameters during the CL process. LoRA methods further introduce low-rank matrices into the model parameters, significantly reducing memory overhead and computational costs associated with fine-tuning. Prompt tuning achieves similar efficiency by learning task-specific prompts, leaving the majority of model weights unchanged. However, current PEFT methods still face challenges in dynamic incremental learning scenarios. For example, InfLoRA~\cite{liang2024inflora} employs rehearsal strategies, which require extensive storage of past task features. SD-LoRA~\cite{wu2025s}, while rehearsal-free, lacks explicit awareness of task similarity, making balance of parameter management across tasks increasingly difficult. Furthermore, the management of multiple adapters or prompts can complicate model deployment and reduce inference efficiency in practical applications.

To overcome these limitations, we propose a novel CL framework called \textbf{Factorized Low-Rank Meta-Prompting (FM-LoRA)}. First, we introduce Factorized Low-Rank Adaptation (F-LoRA) to confine parameter updates into a shared, low-rank subspace. Unlike traditional LoRA methods, F-LoRA explicitly factorizes task-specific updates into shared and task-dependent components, effectively capturing commonalities between tasks while significantly reducing interference, particularly beneficial when tasks share input or output similarities. Then, we propose a Dynamic Rank Selector (DRS) that adaptively determines the rank of the parameter updates based on task complexity. This mechanism incorporates task similarity awareness, which allocates a larger subspace capacity to tasks that differ substantially from previous knowledge and smaller capacities to tasks closely related to previous ones, ensuring efficient resource use and minimizing redundant parameter changes. Finally, we integrate Dynamic Meta-Prompting (DMP), a small set of learnable tokens prepended to each input, to anchor invariant features across tasks. This approach further maintains stable contextual cues, reinforcing shared representations across similar tasks, and effectively preventing the drift of critical learned features over time. Through the synergy of these components, our framework efficiently leverages task similarity, dynamically adapts to varying complexity, and robustly mitigates catastrophic forgetting without any rehearsal of prior task data.

We evaluated our method on a series of benchmark datasets, demonstrating its superior performance in both retaining prior knowledge and adapting to new tasks compared to existing CL techniques. Our contributions include:

\begin{itemize}
    \item We proposed a new rehearsal-free PEFT CL framework with task similarity awareness that leverages F-LoRA with dynamic rank selection and meta-prompting.
    \item Our framework integrates factorized low-rank parameter updates, dynamic adaptation capacity, and task-aware prompting to efficiently handle task similarity without data rehearsal. By explicitly considering task similarity, FM-LoRA effectively reduces interference between tasks, balancing stability and plasticity in incremental learning scenarios.
    \item We conduct comprehensive experiments to demonstrate the effectiveness of our approach in mitigating catastrophic forgetting under both class-incremental (CIL) and domain-incremental (DIL) scenarios. Specifically, we evaluate on ImageNet-R~\cite{hendrycks2021many}, CIFAR100~\cite{krizhevsky2009learning}, and CUB200~\cite{WahCUB_200_2011} for CIL, while using DomainNet~\cite{peng2019moment} for DIL. Our method consistently outperforms state-of-the-art (SOTA) CL approaches across all these widely used benchmarks, confirming its robustness and generalizability.

\end{itemize}

\section{Related Works}
\textbf{Continual Learning (CL).} CL aims to develop models capable of incrementally acquiring new knowledge while preserving previously learned information and generalizing across sequential tasks, thereby addressing the critical challenge known as catastrophic forgetting. Existing CL approaches typically fall into three categories: rehearsal-based methods~\cite{rebuffi2017icarl,chaudhry2018efficient,lopez2017gradient,liang2024inflora}, which maintain a representative subset of data from prior tasks and replay them during subsequent training; regularization-based methods~\cite{kirkpatrick2017overcoming,aljundi2018memory,zenke2017continual,li2017learning,luo2012incremental}, which introduce additional constraints to limit parameter modifications deemed essential for preserving prior knowledge; and architecture-based methods~\cite{fernando2017pathnet,zhou2024expandable,yan2021dynamically,wang2022continual}, which dynamically adjust or expand the model architecture to allocate task-specific parameters, thereby isolating learned knowledge. Among various CL scenarios, this work specifically addresses class-incremental learning (CIL) and domain-incremental learning (DIL), two particularly challenging yet realistic setting in which task and domain identities are unavailable at inference. Traditional CIL and DIL methods often require extensive retraining or tuning from scratch, leading to high computational costs, increased risk of overfitting, and significant interference between tasks. A primary challenge in CL is striking an effective balance between stability and plasticity, ensuring that the model remains stable enough to retain previously acquired knowledge but flexible enough to rapidly adapt to new tasks. 

Although rehearsal-based methods can alleviate forgetting, they require significant memory storage and may raise privacy concerns. Regularization-based approaches, though designed to preserve critical parameters, can impede the model's capacity to integrate novel information. Similarly, architecture-based strategies that expand the capacity of the model often result in unsustainable growth and increased complexity. Recent advances have sought to overcome these limitations by exploring parameter-efficient fine-tuning (PEFT) techniques, which update only a small subset of parameters and leverage the structure of pre-trained models.

\begin{figure*}[t!]
    \centering
    \includegraphics[width=1\linewidth]{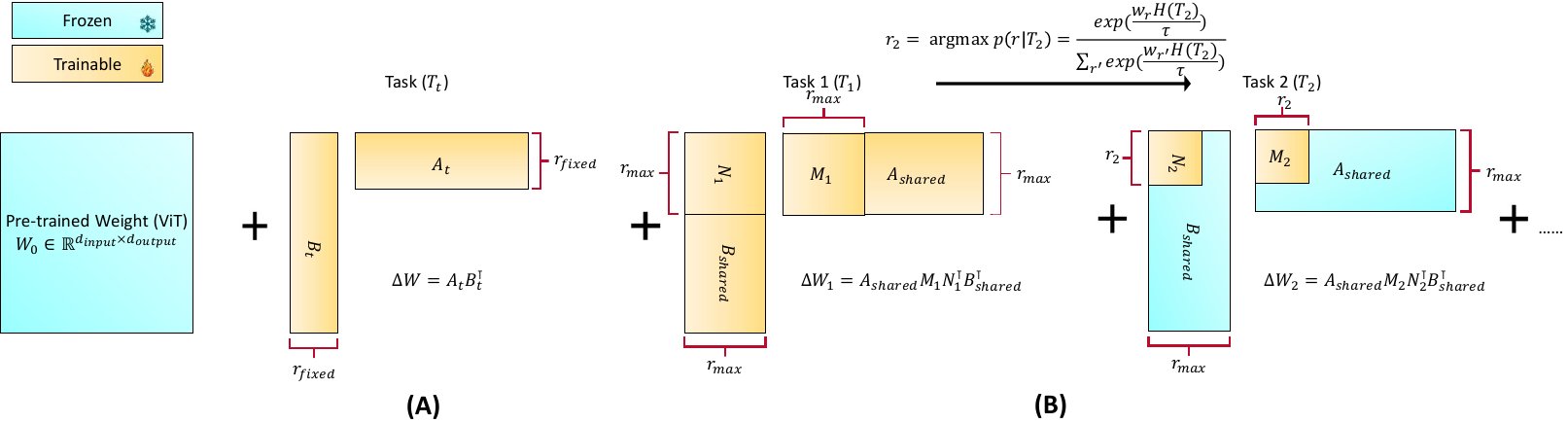}
    \caption{Illustration of incremental weight updates in standard LoRA and our proposed FM-LoRA within CL scenarios. \textbf{(A)}. depicts vanilla low-rank adaptation with {\small$r_{fixed} \ll \min\{d_{input},d_{output}\}$}. \textbf{(B)}. demonstrates FM-LoRA’s learning process across two sequential tasks where this illustrates incremental weight updates in FM-LoRA, showing that task 1 learns maximum-rank shared bases $A_{\text{shared}}$ and $B_{\text{shared}}$, while task 2 dynamically selects a lower rank $r_2$ via DRS, producing updates $\Delta W_t = A_{\text{shared}} M_t N_t^\top B_{\text{shared}}^\top$.}
    \label{fig:fm-lora}
\end{figure*}

\noindent\textbf{Parameter-Efficient Fine-Tuning-based approach in CL}. Parameter-Efficient Fine-Tuning (PEFT)~\cite{ding2023parameter} offers a promising direction for efficiently adapting pre-trained foundation models in CL scenarios. Since complete fine-tuning of large Transformer-based models for each new task is computationally prohibitive, PEFT techniques selectively update only a small subset of parameters. Representative PEFT methods include adapters, which insert compact, task-specific modules within Transformer layers; prompt-tuning~\cite{lester2021power} and prefix-tuning~\cite{li2021prefix}, which introduce learnable input embeddings or prefix vectors to condition pre-trained models; and Low-Rank Adaptation (LoRA)~\cite{hu2022lora}, which adds low-rank parameter updates to pre-trained weights for efficient task adaptation. 

Although PEFT strategies have demonstrated success in single-task or offline multi-task learning, their potential in CL settings remains largely unexplored, motivating further investigation into their application and limits.

\noindent\textbf{CL with Vision Transformers (ViTs).} CL with ViTs has recently gained considerable attention due to the strong generalization capabilities and knowledge transfer potential provided by large pre-trained architectures. Methods such as L2P~\cite{wang2022learning}, DualPrompt~\cite{wang2022dualprompt}, and CODA-Prompt~\cite{smith2023coda} effectively leverage ViTs combined with prompt-tuning strategies to mitigate catastrophic forgetting. More recent developments, such as Hide-Prompt~\cite{wang2024hierarchical} and InfLoRA~\cite{liang2024inflora}, further improve performance through extensive sample storage, though at increased storage cost. Despite these advancements, existing CL methods that utilize foundation models fail to simultaneously achieve all desirable properties: parameter efficiency, minimal reliance on rehearsal, and robustness against catastrophic forgetting. To bridge this gap, SD-LoRA~\cite{wu2025s} was introduced as a rehearsal-free LoRA-based PEFT approach explicitly designed for CL scenarios. SD-LoRA incorporates recent model merging techniques, positioning it as complementary but distinct within ongoing CL research. 

However, SD-LoRA exhibits certain limitations, such as potential challenges in maintaining performance across a growing number of tasks, particularly due to its lack of explicit task similarity awareness. To address these weaknesses, we propose a new method, called FM-LoRA, designed to enhance CL performance with ViTs by integrating parameter-efficient fine-tuning techniques tailored for CL scenarios.

\section{Methodology}

\subsection{Problem Setting}

We consider a \emph{rehearsal-free CL} scenario, where a ViT model sequentially encounters a set of $T$ tasks, denoted as $\{\mathcal{T}_1,\dots,\mathcal{T}_T\}$. Each task $\mathcal{T}_t$ provides a dataset $\mathcal{D}_t = \{(x_i,y_i)\}$, where $x_i$ denotes input embeddings and $y_i$ are labels. Initially, the model is pre-trained with weight matrices $\{W_0\}$ (such as Query, Key, Value, or MLP layers), each of dimension $d_{\text{input}}\times d_{\text{output}}$. In this incremental learning setup, the goal is to adapt the model parameters to each new dataset $\mathcal{D}_t$ without revisiting old data or causing catastrophic forgetting. To achieve this, we propose a novel incremental learning framework leveraging three components: \emph{Factorized Low-Rank Adaptation (F-LoRA)}, a \emph{Dynamic Rank Selector (DRS)}, and a \emph{Dynamic Meta-Prompting (DMP)} module, each carefully designed to preserve existing knowledge while effectively acquiring new information.

\subsection{Factorized Low-Rank Adaptation}
To efficiently adapt to new tasks while mitigating catastrophic forgetting, we propose \textbf{Factorized Low-Rank Adaptation (F-LoRA)}, a novel low-dimensional reparameterization of incremental weight updates. Consider a standard weight matrix $W \in \mathbb{R}^{d_{\text{input}} \times d_{\text{output}}}$, where $d_{\text{input}}$ and $d_{\text{output}}$ denote input and output dimensions, respectively. Rather than directly fine-tuning $W$, we introduce a parameter-efficient incremental update $\Delta W$, expressed as:
\begin{equation}
W' = W + \Delta W.
\end{equation}

In standard LoRA methods, this incremental update is represented as the product of two fully task-specific matrices $\Delta W_t = A_t B_t^\top$, with $A_t \in \mathbb{R}^{d_{\text{input}} \times r}$ and $B_t \in \mathbb{R}^{d_{\text{output}} \times r}$. Thus, the count of parameters per task is $r(d_{\text{input}} + d_{\text{output}})$, often resulting in unnecessary redundancy and higher interference among tasks.

In contrast, our F-LoRA explicitly factorizes the incremental update into shared and task-specific components:
\begin{equation}
\Delta W_t = A_{\text{shared}} M_t N_t^\top B_{\text{shared}}^\top
\end{equation}
where $A_{\text{shared}} \in \mathbb{R}^{d_{\text{input}}\times r}$ and $B_{\text{shared}} \in \mathbb{R}^{d_{\text{output}}\times r}$ represent global low-rank bases learned across tasks, capturing reusable directions common among multiple incremental adaptations. The matrices $M_t, N_t \in \mathbb{R}^{r\times r}$ are task-specific coefficients that provide targeted adjustments to these shared directions for the task $\mathcal{T}_t$. Consequently, F-LoRA dramatically reduces the overhead per task parameter from $r(d_{\text{input}} + d_{\text{output}})$ in standard LoRA to just $2r^2$, with globally shared bases incurring a one-time cost of $r(d_{\text{input}} + d_{\text{output}})$ parameters (Figure~\ref{fig:fm-lora}).

This structured factorization confines parameter updates to a stable, low-dimensional manifold explicitly defined by fixed bases $A_{\text{shared}}$ and $B_{\text{shared}}$. Specifically, we first select a maximum candidate rank $r_{\text{max}}$, learn global bases $A_{\text{shared}} \in \mathbb{R}^{d_{\text{input}}\times r_{\text{max}}}$ and $B_{\text{shared}} \in \mathbb{R}^{d_{\text{output}}\times r_{\text{max}}}$ during the initial incremental task, and subsequently freeze these matrices. For each subsequent incremental task, the DRS adaptively chooses a task-specific rank $r_t \leq r_{\text{max}}$ by selecting a subset of columns from these fixed bases. Consequently, for each incremental update $\Delta W_t$, the column and row spaces satisfy:
\begin{align}
\text{colspace}(\Delta W_t) &\subseteq \text{colspace}(A_{\text{shared}}[:, :r_t]), \\[4pt]
\text{rowspace}(\Delta W_t) &\subseteq \text{rowspace}(B_{\text{shared}}[:, :r_t]).
\end{align}
ensuring updates lie within a stable and controlled subspace. This constraint inherently reduces subspace overlap between task-specific updates, significantly limiting potential interference and catastrophic forgetting.

To further quantify interference, we define it using the Frobenius inner product between task updates:
\begin{align}
\text{Interference}(\mathcal{T}_i, \mathcal{T}_j) & = \langle \Delta W_i, \Delta W_j \rangle_F \\\nonumber
 & = \text{Tr}(\Delta W_i^\top \Delta W_j).
\end{align}
Substituting the F-LoRA parameterization yields as
\begin{align}
&\langle \Delta W_i, \Delta W_j \rangle_F =\\\nonumber 
&\text{Tr}(B_{\text{shared}} N_i M_i^\top A_{\text{shared}}^\top A_{\text{shared}} M_j N_j^\top B_{\text{shared}}^\top).
\end{align}
When $A_{\text{shared}}$ and $B_{\text{shared}}$ are orthonormal (i.e., $A_{\text{shared}}^\top A_{\text{shared}} = I_r$, $B_{\text{shared}}^\top B_{\text{shared}} = I_r$), this interference measure simplifies significantly:
\begin{equation}
\langle \Delta W_i, \Delta W_j \rangle_F = \text{Tr}(N_i M_i^\top M_j N_j^\top).
\end{equation}

\begin{figure}[t!]
    \centering
         \includegraphics[width=1\linewidth]{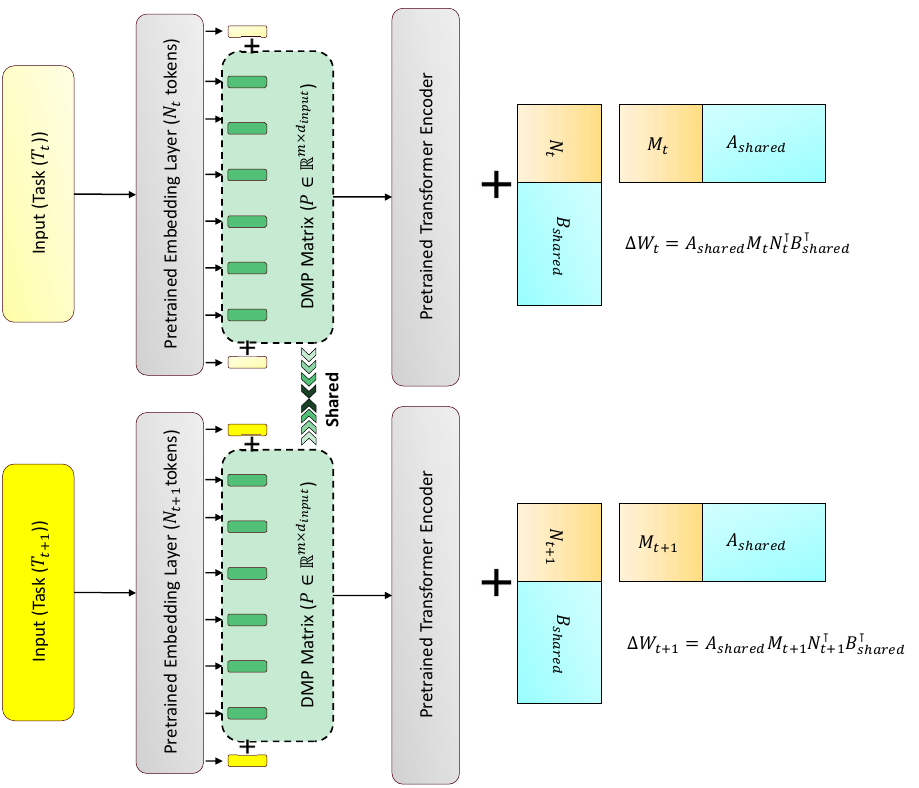}
    \caption{ An illustration of the proposed dynamic meta prompting (DMP), where a learnable prompt matrix is prepended to each input sequence. This design stabilizes representations across incremental tasks by providing a shared context throughout training.} 
\label{fig:dmp}
\end{figure}

In particular, interference now depends solely on the low-dimensional task-specific matrices ($M_t, N_t$) in a $r^2$-dimensional space. In comparison, standard LoRA interference depends directly on high-dimensional matrices $A_t, B_t$, scaling with input and output dimensions, thus increasing the potential for substantial task interference.

Gradient-based optimization in F-LoRA further reinforces this structured constraint. Specifically, gradients with respect to the task-specific coefficients are computed via:
\begin{align}
\frac{\partial L}{\partial M_t} &= A_{\text{shared}}^\top \frac{\partial L}{\partial \Delta W_t} B_{\text{shared}} N_t, \\[4pt]
\frac{\partial L}{\partial N_t} &= B_{\text{shared}}^\top \left(\frac{\partial L}{\partial \Delta W_t}\right)^\top A_{\text{shared}} M_t.
\end{align}

These gradient computations inherently confine optimization steps within the predefined subspaces spanned by the global bases, providing theoretical guarantees that task-specific updates will not arbitrarily disrupt previously learned representations.

Furthermore, the low-dimensional manifold defined by F-LoRA updates:
\begin{align}
&\mathcal{M}_{\text{F-LoRA}} =\\\nonumber
&\{\Delta W = A_{\text{shared}} M N^\top B_{\text{shared}}^\top \mid M,N \in \mathbb{R}^{r\times r}\}.
\end{align}
has dimension $2r^2$, significantly lower than the $r(d_{\text{input}} + d_{\text{output}} - r)$ dimension associated with standard LoRA updates. This dimensional reduction is beneficial from the manifold learning perspective, promoting updates that generalize better due to reduced overfitting and controlled update trajectories.

In summary, our proposed F-LoRA demonstrates significant advantages with dramatic reductions in per-task parameter overhead, rigorous confinement of updates to low-dimensional and interference-resistant subspaces, and providing stronger stability guarantees for incremental updates, collectively achieving a robust balance between model stability and adaptability in CL scenarios.

\subsection{Dynamic Rank Selector}
While F-LoRA confines parameter updates within rank-$r$ subspaces, using a fixed rank across tasks may not optimally accommodate varying task complexities, potentially limiting model adaptability and performance. To rigorously address this issue, we introduce a \textbf{dynamic rank selector (DRS)}, which adaptively determines an effective rank $r_t$ based on a quantifiable complexity measure $H(\mathcal{T}_t)$ computed individually for each task. Specifically, at the beginning of training each incremental task $\mathcal{T}_t$, we perform a brief preliminary training epoch to estimate the validation loss, subsequently using this loss as the task complexity measure $H(\mathcal{T}_t)$ (Figure~\ref{fig:fm-lora}).

Let $\{w_r\}$ be learnable weights associated with each candidate rank $r$. The probability distribution over candidate ranks is modeled via a differentiable Gumbel-Softmax mechanism:
\begin{equation}
p(r|\mathcal{T}_t) = \frac{\exp\left(\frac{w_r\,H(\mathcal{T}_t)}{\tau}\right)}{\sum_{r'} \exp\left(\frac{w_{r'}\,H(\mathcal{T}_t)}{\tau}\right)}.
\end{equation}
where $\tau$ is a temperature hyperparameter controlling distribution sharpness, and $r'$ is an indexing variable iterating over all candidate ranks. From this distribution, the effective rank for task $\mathcal{T}_t$ is selected as:
\begin{equation}
r_t = \arg\max_r\, p(r|\mathcal{T}_t).
\end{equation}

Intuitively, tasks exhibiting higher complexity, which are indicated by a higher initial validation loss, result in the selection of a larger rank $r_t$, thus enabling more expressive adaptation. In contrast, simpler tasks yield lower complexity scores and result in the selection of smaller ranks, ensuring parameter-efficient updates. Moreover, this mechanism implicitly incorporates the awareness of task similarity. For instance, when a new task is closely related to previous tasks, its complexity measure tends to be lower, prompting the selection of a smaller rank, and thereby reducing redundant updates, and if tasks differ substantially, a higher rank is selected to capture the novel information. In this way, the DRS explicitly balances model flexibility and stability, mitigating catastrophic forgetting by precisely aligning parameter capacity with task-specific demands and inherent task similarity.

\subsection{Dynamic Meta-Prompt}
Although F-LoRA and DRS localize and adapt updates efficiently, incremental learning without rehearsal inherently risks drift in learned representations. We address this by introducing a learnable \textbf{dynamic meta-prompt (DMP)} matrix $P\in\mathbb{R}^{m\times d}$, consisting of $m$ prompt tokens with embedding dimension $d$ that stabilize invariant knowledge across tasks.

Given input embeddings $\{\mathbf{x}_1,\dots,\mathbf{x}_N\}\subset\mathbb{R}^d$, we augment the input sequence by prepending the meta-prompt tokens:
\begin{equation}
\mathbf{z}_i = 
\begin{cases}
\mathbf{p}_i, & 1\le i\le m.\\[5pt]
\mathbf{x}_{i-m}, & m+1\le i\le m+N.
\end{cases}
\end{equation}
with each $\mathbf{p}_i$ being the $i$-th row of the prompt matrix $P$ and $N$ represents the number of input embedding tokens.

The transformer backbone, parameterized by $\theta$ (inclusive of F-LoRA parameters), processes these augmented inputs to produce output embeddings $\mathbf{h}_{\text{out}}\in\mathbb{R}^{(m+N)\times d}$. We then optimize task-specific performance via standard supervised loss:
\begin{equation}
L_{\text{sup}}(\theta,P;\mathcal{D}_t) = \sum_{(x_i,y_i)\in\mathcal{D}_t}\ell\left(\mathrm{ViT}(\{\mathbf{P};x_i\};\theta),y_i\right).
\end{equation}
where $\ell$ denotes the loss function used to measure the discrepancy between the model's prediction and the ground-truth label, and $\{\mathbf{P};x_i\}$ denotes concatenated prompt and input embeddings.

Our DMP provides a stable context vector set that reduces parameter drift in subsequent adaptations. From the perspective of optimization dynamics, consider the gradient update:
\begin{equation}
\frac{\partial L_{\text{sup}}}{\partial \mathbf{P}} = \sum_{(x_i,y_i)\in\mathcal{D}_t} \frac{\partial \ell}{\partial \mathbf{h}_{\text{out}}}\frac{\partial \mathbf{h}_{\text{out}}}{\partial \mathbf{P}}.
\end{equation}

Since $P$ updates are influenced by multiple tasks due to the shared nature of this DMP, the prompt matrix naturally encodes a representation subspace invariant to task-specific changes. Thus, it acts as implicit memory, constraining parameter drift and ensuring that the learned representations remain closely aligned with the stable subspace defined by F-LoRA and the dynamically allocated capacity determined by DRS throughout sequential learning.

By systematically integrating F-LoRA, DRS and DMP, our proposed FM-LoRA offers a task-aware CL framework. It explicitly reduces interference through structured, controlled low-rank adaptation, dynamically matches parameter-update capacity to task complexity and similarity, and further stabilizes invariant representations through implicit memory provided by the DMP. Collectively, these components rigorously address the stability-plasticity trade-offs inherent in CL scenarios, reducing catastrophic forgetting, and facilitating efficient CL without rehearsal.

\begin{table*}[t]
 
\begin{center}
\scalebox{0.9}{ 
\begin{tabular}{c|cc|cc|cc}
\toprule
 \multicolumn{1}{l|}{\multirow{2}{*}{Method}} & \multicolumn{2}{c|}{ImageNet-R ($N=5$)}  & \multicolumn{2}{c|}{ImageNet-R ($N=10$)} & \multicolumn{2}{c}{ImageNet-R ($N=20$)}  \\ \cline{2-7} 
        &Acc $\uparrow$  & AAA $\uparrow$ & Acc $\uparrow$  & AAA $\uparrow$   & Acc $\uparrow$ & AAA $\uparrow$ \\
\midrule
\multicolumn{1}{l|}{\multirow{1}{*}{Fine-Tuning}} & 64.92$\pm$0.87 & 75.57$\pm$0.50 & 60.57$\pm$1.06 & 72.31$\pm$1.09 & 49.95$\pm$1.31 & 65.32$\pm$0.84 \\
\hline
\multicolumn{1}{l|}{\multirow{1}{*}{L2P~\cite{wang2022learning}}} &  73.04$\pm$0.71 & 76.94$\pm$0.41 & 71.26$\pm$0.44 & 76.13$\pm$0.46 & 68.97$\pm$0.51 & 74.16$\pm$0.32 \\

\multicolumn{1}{l|}{\multirow{1}{*}{Dual-Prompt~\cite{wang2022dualprompt}}} & 69.99$\pm$0.57 & 72.24$\pm$0.41 & 68.22$\pm$0.20 & 73.81$\pm$0.39 & 65.23$\pm$0.45 & 71.30$\pm$0.16 \\

\multicolumn{1}{l|}{\multirow{1}{*}{Coda-Prompt~\cite{smith2023coda}}} &  76.63$\pm$0.27 & 80.30$\pm$0.28 & 74.05$\pm$0.41 & 78.14$\pm$0.39 & 69.38$\pm$0.33 & 73.95$\pm$0.63 \\

\multicolumn{1}{l|}{\multirow{1}{*}{Hide-Prompt~\cite{wang2024hierarchical}}} & 74.77$\pm$0.25 & 78.15$\pm$0.24 & 74.65$\pm$0.14 & 78.46$\pm$0.18 & 73.59$\pm$0.19 & 77.93$\pm$0.19 \\

\multicolumn{1}{l|}{\multirow{1}{*}{C-LoRA~\cite{smith2023continual}}} & 75.85$\pm$0.31 & 79.17$\pm$0.28 & 71.89$\pm$0.51 & 76.16$\pm$0.61 & 65.71$\pm$0.53 & 71.66$\pm$0.49 \\

\multicolumn{1}{l|}{\multirow{1}{*}{InfLoRA~\cite{liang2024inflora}}} & 76.95$\pm$0.23 & 81.81$\pm$0.14 & 74.75$\pm$0.64 & 80.67$\pm$0.55 & 69.89$\pm$0.56 & 76.68$\pm$0.57 \\

\multicolumn{1}{l|}{\multirow{1}{*}{SD-LoRA~\cite{wu2025s}}} & \textbf{79.15}$\pm$0.20 & \textbf{83.01}$\pm$0.42 &\underline{77.34}$\pm$0.35 & \underline{82.04}$\pm$0.24 & \underline{75.26}$\pm$0.37 & \underline{80.22}$\pm$0.72 \\
\cline{1-7}
\multicolumn{1}{l|}{\multirow{1}{*}{\textbf{FM-LoRA}}} & \underline{78.93}$\pm$0.38 & \underline{82.37}$\pm$0.45 & \textbf{77.88}$\pm$0.21 & \textbf{82.59}$\pm$0.31  & \textbf{76.47}$\pm$0.65 & \textbf{80.97}$\pm$0.35 \\ 

\multicolumn{1}{l|}{\multirow{1}{*}{\textbf{FM-LoRA (Without DMP)}}} & 77.32$\pm$0.33 & 81.73$\pm$0.42 & 76.96$\pm$0.61 & 81.83$\pm$0.23  & 75.12$\pm$0.75 & 80.18$\pm$0.83 \\ 
\bottomrule
\end{tabular}
}
\end{center}
\caption{Model performance comparisons and ablation studies on ImageNet-R across different task lengths, including standard deviation. All reported values are the mean of five runs. The best and second-best results for each dataset and task are highlighted in {\textbf{bold}} and \underline{underline}, respectively. All results from FM-LoRA used the optimal $m$ values for DMP.}
\label{exp:tab-inr}
\end{table*}

\section{Experiments}

\subsection{Implementation Deatils}

To thoroughly validate the effectiveness of our proposed FM-LoRA, we conduct experiments on several widely used CL benchmarks thoroughly assess the effectiveness of our framework. Specifically, we conduct experiments on standard CL benchmarks: ImageNet-R~\citep{hendrycks2021many}, CIFAR100~\cite{krizhevsky2009learning},  and CUB200~\cite{WahCUB_200_2011} for CIL and DomainNet~\cite{peng2019moment} for DIL. Following recent practices~\cite{liang2024inflora,wu2025s}, we incrementally partition ImageNet-R, which contains 200 ImageNet classes~\cite{deng2009imagenet} into 5, 10, and 20 tasks (with 40, 20, and 10 classes per task, respectively). CIFAR-100 is a widely used dataset for image classification, comprising 60,000 images evenly distributed across 100 classes (600 images per class). For our experiments, we partition CIFAR-100 into 10 tasks, with each task containing 10 classes. Similarly, CUB-200 is a fine-grained dataset for bird classification that consists of 11,788 images spanning 200 classes; we divide this dataset into 10 tasks, with each task encompassing 20 species. We partition DomainNet, a large scale benchmark containing over 600,000 images spanning 345 categories across six distinct domains (Clipart, Painting, Real, Sketch, Infograph, and Quickdraw) that exhibit substantial domain shifts into 5 tasks of 69 categories each. In our experiments, each dataset was partitioned into training, validation, and test sets following standard protocols. For example, for ImageNet-R, we allocate 70\% of the samples for training, 10\% for validation, and 20\% for testing within each incremental task; similar splits were applied to CIFAR100, CUB200, and DomainNet.

\begin{table*}[t!]
 
\begin{center}
\scalebox{0.9}{ 
\begin{tabular}{c|cc|cc|cc}
\toprule
 \multicolumn{1}{l|}{\multirow{2}{*}{Method}} & \multicolumn{2}{c|}{CIFAR100 ($N=10$)}  & \multicolumn{2}{c|}{CUB200 ($N=10$)} & \multicolumn{2}{c}{DomainNet ($N=5$)}  \\ \cline{2-7} 
        &Acc $\uparrow$  & AAA $\uparrow$ & Acc $\uparrow$  & AAA $\uparrow$   & Acc $\uparrow$ & AAA $\uparrow$ \\
\midrule
\multicolumn{1}{l|}{\multirow{1}{*}{Fine-Tuning}} & 69.49$\pm$0.50 & 80.35$\pm$0.87  & 51.43$\pm$1.41 & 69.74$\pm$0.93 & 51.46$\pm$0.47 & 67.08$\pm$1.13 \\
\hline
\multicolumn{1}{l|}{\multirow{1}{*}{L2P~\cite{wang2022learning}}} &  83.18$\pm$1.20 & 87.69$\pm$1.05 & 65.18$\pm$2.49 & 76.12$\pm$1.27 & 70.26$\pm$0.25 & 75.83$\pm$0.98 \\

\multicolumn{1}{l|}{\multirow{1}{*}{Dual-Prompt~\cite{wang2022dualprompt}}} & 81.48$\pm$0.86 & 86.41$\pm$0.66 & 68.00$\pm$1.06 & 79.40$\pm$0.88 & 68.26$\pm$0.90 & 73.84$\pm$0.45 \\

\multicolumn{1}{l|}{\multirow{1}{*}{Coda-Prompt~\cite{smith2023coda}}} & 86.31$\pm$0.12 & 90.67$\pm$0.22 & 71.92$\pm$0.33 & 78.76$\pm$0.65 & 70.58$\pm$0.53 & 76.68$\pm$0.44 \\

\multicolumn{1}{l|}{\multirow{1}{*}{Hide-Prompt~\cite{wang2024hierarchical}}} & 86.45$\pm$0.04 & 91.34$\pm$0.13 & 74.77$\pm$0.25 & 78.15$\pm$0.24 & 72.20$\pm$0.08 & 77.01$\pm$0.04 \\

\multicolumn{1}{l|}{\multirow{1}{*}{C-LoRA~\cite{smith2023continual}}} & 82.97$\pm$0.47 & 87.11$\pm$0.25 & 69.99$\pm$0.61 & 79.11$\pm$0.69 & 69.34$\pm$0.47 & 74.66$\pm$0.51 \\

\multicolumn{1}{l|}{\multirow{1}{*}{InfLoRA~\cite{liang2024inflora}}} & 86.75$\pm$0.35 & 91.72$\pm$0.15 & 70.82$\pm$0.23 & 81.39$\pm$0.14 & 71.59$\pm$0.23 & 78.29$\pm$0.50 \\

\multicolumn{1}{l|}{\multirow{1}{*}{SD-LoRA~\cite{wu2025s}}} & 88.01$\pm$0.31 & 92.54$\pm$0.18 & 77.48$\pm$0.20 & 85.59$\pm$0.44 & \textbf{72.82}$\pm$0.37 & \textbf{78.89}$\pm$0.72 \\
\cline{1-7}
\multicolumn{1}{l|}{\multirow{1}{*}{\textbf{FM-LoRA}}} & \textbf{88.37}$\pm$0.29 & \textbf{92.66}$\pm$0.14 & \textbf{77.79}$\pm$0.22 & \textbf{86.01}$\pm$0.39  & \underline{72.66}$\pm$0.48 & \underline{78.31}$\pm$0.85 \\ 

\multicolumn{1}{l|}{\multirow{1}{*}{\textbf{FM-LoRA (Without DMP)}}} & 87.37$\pm$0.41 & 92.03$\pm$0.29 & 77.08$\pm$0.25 & 84.89$\pm$0.42  & 71.19$\pm$0.85 & 77.16$\pm$0.92 \\ 
\bottomrule
\end{tabular}
}
\end{center}
\caption{Model performance comparisons and ablation studies on CIFAR100, CUB200, and DomainNet across different task lengths, including standard deviation. All reported values are the mean of five runs.}
\label{exp:tab-inr2}
\end{table*}

We utilize a Vision Transformer (ViT-B/16)~\cite{dosovitskiy2020image} pretrained on ImageNet as the backbone due to its proven efficacy in CL contexts. Baseline comparisons include standard fine-tuning, LoRA~\cite{hu2022lora}, adapters~\cite{houlsby2019parameter}, as well as recent PEFT-based CL methods, including SD-LoRA and InfLoRA~\cite{liang2024inflora,wu2025s}.

To enhance adaptability and efficiency, we introduce a \textbf{DRS}, an adaptive mechanism to automatically determine the optimal rank \( r \) per incremental task. Specifically, candidate ranks \( r \in \{4, 8, 16, 32\} \) are evaluated based on validation set performance, selecting the minimal rank that achieves competitive accuracy, thus dynamically balancing complexity and performance. The shared basis matrices \( A_{\text{shared}} \) and \( B_{\text{shared}} \) are initialized randomly and jointly optimized only during the initial incremental task, remaining fixed afterward. Task-specific matrices \( M_t, N_t \) are learned individually for each task based on the dynamically chosen rank.  Additionally, for the DMP module, the prompt matrix $P \in \mathbb{R}^{m \times d_\text{input}}$ is used to prepend a fixed set of learnable tokens to each input sequence. We initially set $m=10$, and keep investigating the optimal values for different lengths of tasks, further providing a balanced trade-off between model expressiveness and parameter efficiency in all of our experiments.

\begin{figure}[t!]
    \centering
         \includegraphics[width=1\linewidth]{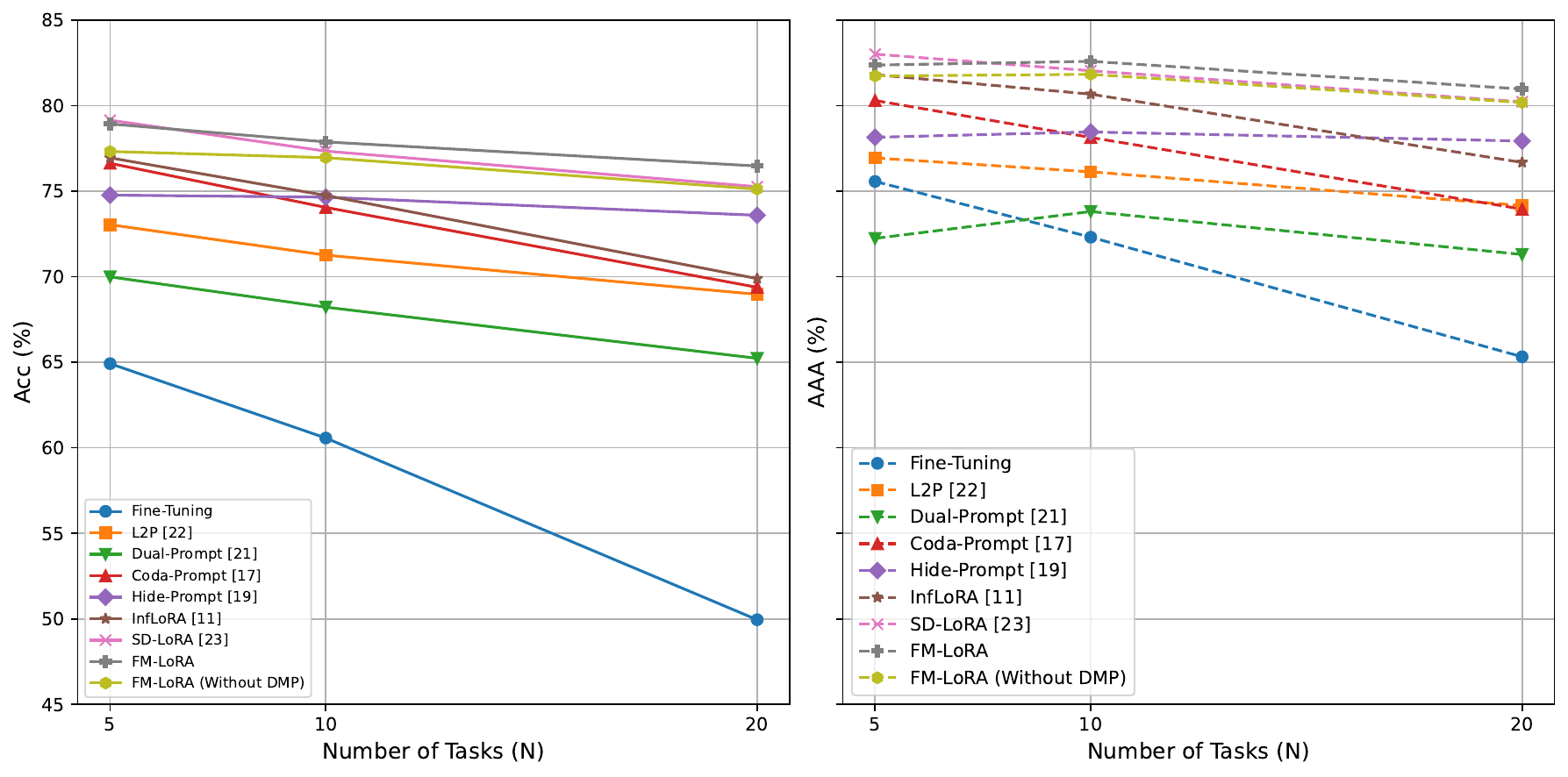}
    \caption{Performance trajectory comparisons (\textbf{Acc} and\textbf{AAA}, the lower the better) across varying numbers of sequential tasks ($N$=5, 10, 20) on ImageNet-R for different CL methods.}\vspace{-10pt}
\label{fig:imag_performance}
\end{figure}

All models are optimized with AdamW~\cite{loshchilov2017decoupled}, using a learning rate of \(2\times10^{-4}\), a cosine annealing scheduler, and a weight decay of \(0.05\). Training spans 30 epochs per task with a batch size of 256, distributed over two NVIDIA A100 GPUs. Consistent pre-processing and data augmentation ensure fair comparisons.

We employ two primary evaluation metrics to comprehensively measure CL performance. The final average precision (\textbf{Acc}) reflects the overall performance of all \( N \) incremental tasks after completing the entire learning sequence, providing insight into the final knowledge retained. In contrast, the average anytime accuracy (\textbf{AAA}) captures the average performance of the model throughout the sequential learning process, effectively reflecting how consistently the model maintains and accumulates knowledge during incremental training.

\subsection{Comparison with SOTA}

\subsubsection{Performance on ImageNet-R with different task lengths.}
We present model performance comparisons on ImageNet-R across different task lengths ($N=5,10,20$) in Table~\ref{exp:tab-inr}. Our proposed \textbf{FM-LoRA} consistently achieves the highest final accuracy (\textit{Acc}) and average anytime accuracy (\textit{AAA}) for longer task lengths ($N=10, 20$), surpassing strong baselines such as InfLoRA and SD-LoRA. In particular, although FM-LoRA does not outperform SD-LoRA in the 5-task setting, its performance advantage becomes increasingly pronounced as the number of tasks grows. As task length increases from $N=5$ to $N=20$, the \textit{Acc} and \textit{AAA} for most methods degrade significantly due to accumulated interference and catastrophic forgetting. However, FM-LoRA maintains superior margins (Figure~\ref{fig:imag_performance}). For instance, in the 20-task setting, FM-LoRA outperforms SD-LoRA by approximately 1.21\% in \textit{Acc} and 0.75\% in \textit{AAA}, and exceeds InfLoRA by even larger margins.

This enhanced performance for longer task lengths is attributed to FM-LoRA’s implicit task similarity awareness enabled by the integration of our DRS with F-LoRA. Our DRS dynamically adjusts the effective rank based on a quantifiable measure of task complexity, allocating greater capacity to tasks that are more distinct from previous ones. As a result, FM-LoRA not only mitigates catastrophic forgetting but also adapts more effectively to a broader range of task complexities in scenarios with an increasing number of tasks. We also observed that while FM-LoRA without DMP already achieves robust performance, incorporating DMP further enhances stability and boosts accuracy by providing a dedicated implicit memory that mitigates catastrophic forgetting.

\begin{table}[t]
\centering
\resizebox{0.7\columnwidth}{!}{ 
\begin{tabular}{l|c|c}
\hline
\textbf{Method} & \multicolumn{2}{c}{ImageNet-R ($N=10$)}   \\
\cline{1-3}
F-LoRA &Acc $\uparrow$  & AAA $\uparrow$  \\
\hline
+ r = $4$ & 74.23$\pm$0.72  & 78.10$\pm$0.58  \\
+ r = $8$ & \underline{74.68}$\pm$0.68 & \underline{80.31}$\pm$0.42  \\
+ r = $16$ & 74.41$\pm$0.52 & 79.75$\pm$0.31 \\
+ r = $32$ & 73.13$\pm$0.82 & 79.98$\pm$0.77 \\
\hline
+ DRS & \textbf{77.88}$\pm$0.21 & \textbf{82.59}$\pm$0.31  \\
\bottomrule
\end{tabular}
} 
\caption{Performance Comparison on ImageNet-R ($N=10$). We compared  (\textit{Acc}) and  (\textit{AAA}) for F-LoRA with different fixed ranks ($r=4, 8, 16, 32$) as well as the DRS. We demonstrated that the F-LoRA with DRS consistently outperformed F-LoRA with a fixed rank, showing that the adaptive rank selection mechanism more effectively allocates capacity according to task complexity and similarity, thereby reducing catastrophic forgetting and enhancing overall performance in CL scenarios.} 
\label{table:ab}
\end{table}

\begin{figure*}[t!]
    \centering
         \includegraphics[width=0.7\linewidth]{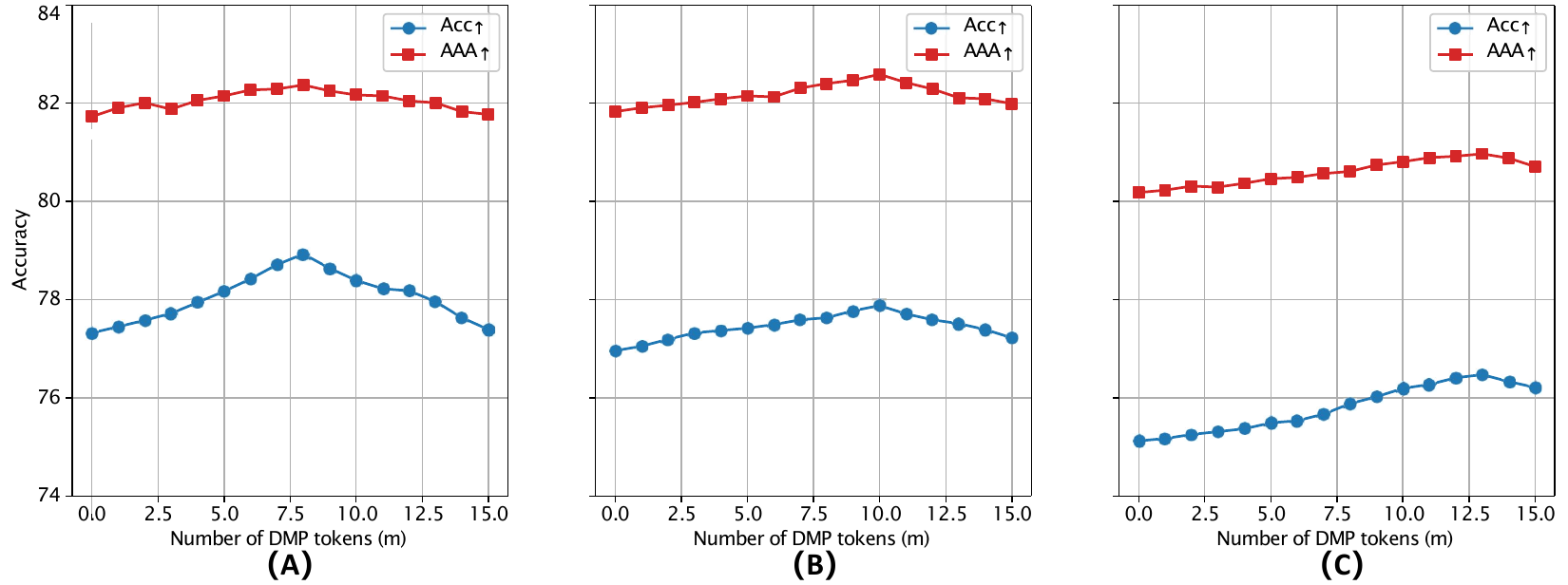}
    \caption{ Detailed ablation study of the proposed DMP with varying initial numbers of shared tokens. \textbf{(A-C)} Results on ImageNet-R with $N=5, 10, 20$ across different $m$ values, demonstrating the importance of adapting DMP complexity based on the number of tasks.} \vspace{-10pt}
\label{fig:dmp_performance}
\end{figure*}

\subsubsection{Performance on more datasets.}
We also summarize the experimental results on three additional benchmarks, each of which poses distinct challenges due to the varying number of classes and domains (Table~\ref{exp:tab-inr2}). Consistent with our previous observations, \textbf{FM-LoRA} maintains superior final accuracy (\textit{Acc}) and average anytime accuracy (\textit{AAA}), underscoring its robustness in more diverse or domain-shifted scenarios. In particular, on CIFAR100 (with 10 tasks) and CUB200 (also with 10 tasks), FM-LoRA shows clear improvements over strong baselines like InfLoRA and SD-LoRA, further validating its ability to learn and retain knowledge across incremental tasks. 

Notably, DomainNet introduces an additional layer of complexity by encompassing six distinct domains in a single dataset. Despite this increased difficulty, FM-LoRA consistently achieves substantial gains over competing methods in both \textit{Acc} and \textit{AAA}, highlighting its strong capacity to handle domain shifts. These results confirm that the synergy of F-LoRA, DRS, and DMP for FM-LoRA design extends effectively beyond ImageNet-R, yielding stable performance across a variety of incremental learning benchmarks and data distributions.

Overall, these findings confirm that FM-LoRA effectively balances plasticity and stability, achieving superior CL performance across multiple datasets and task configurations.

\subsection{Ablation Study}

\noindent\textbf{Impact of DRS}. We investigate the impact of using a single, fixed rank versus dynamically adapting the rank per task. Each fixed-rank configuration strikes a different balance between \textit{Acc} and \textit{AAA} (Table.~\ref{table:ab}), yet none surpasses the overall performance of DRS. When $r$ is too low (for example, $r=4$), the model underfits more complex tasks, leading to lower accuracy. Conversely, a high fixed rank (e.g., $r=32$) risks over-parameterization for simpler tasks, thereby reducing average anytime accuracy and potentially exacerbating forgetting. 

In contrast, our DRS adaptively selects the rank based on each task’s complexity and similarity, allocating sufficient capacity for challenging tasks while avoiding unnecessary overhead on simpler ones. This dynamic strategy not only preserves high accuracy (\textit{Acc}) but also significantly improves \textit{AAA}, indicating better retention of knowledge across tasks. We have observed DRS consistently outperforms any fixed-rank setup in both \textit{Acc} and \textit{AAA}. This highlights how a dynamic rank approach effectively balances model capacity and parameter efficiency, offering superior stability and plasticity for CL scenarios.

\noindent\textbf{Impact of DMP}. We further investigated our proposed DMP, and our ablation study on ImageNet-R demonstrates DMP's effectiveness across different configurations of prompt tokens ($m$) (Figure.~\ref{fig:dmp_performance}). These results reveal that for the DMP module, a lower \(m\) means that fewer prompt tokens are used to form the DMP. While this reduces the model’s capacity to capture and convey invariant contextual information across tasks, potentially limiting the stabilizing effect on learned representations, it also offers improved parameter efficiency by lowering computational and memory overhead. Hence, selecting a proper \(m\) involves a trade-off between efficiency and the richness of the contextual cues provided by the DMP.

Moreover, the optimal choice of \(m\) depends on the length of the task sequence. For shorter task sequences (e.g., \(N=5\)), a lower \(m\) is sufficent because the model faces less task variability and a lower risk of catastrophic forgetting. However, in longer task sequences (e.g., \(N=20\)), the model must handle a broader range of tasks and greater cumulative interference. In these scenarios, a larger \(m\) can supply richer invariant contextual cues as a stronger implicit memory to stabilize learned representations across tasks. 

\section{Conclusion}
In this work, we introduced \textbf{FM-LoRA}, a novel framework that integrates Factorized Low-Rank Adaptation (F-LoRA) with a dynamic rank selector (DRS) and, optionally, a dynamic meta-prompting (DMP) component. By confining incremental weight updates to a stable, low-dimensional subspace, F-LoRA minimizes task interference, while the DRS adaptively adjusts model capacity based on task complexity and similarity, improving flexibility, and reducing overfitting with shared DMP tokens. Our experiments on multiple benchmarks, show that FM-LoRA achieves SOTA performance in both final accuracy and average anytime accuracy, effectively mitigating catastrophic forgetting over longer task sequences.

{
    \small
    \bibliographystyle{ieeenat_fullname}
    \bibliography{main}

\begin{thebibliography}{32}
\providecommand{\natexlab}[1]{#1}
\providecommand{\url}[1]{\texttt{#1}}
\expandafter\ifx\csname urlstyle\endcsname\relax
  \providecommand{\doi}[1]{doi: #1}\else
  \providecommand{\doi}{doi: \begingroup \urlstyle{rm}\Url}\fi

\bibitem[Aljundi et~al.(2018)Aljundi, Babiloni, Elhoseiny, Rohrbach, and Tuytelaars]{aljundi2018memory}
Rahaf Aljundi, Francesca Babiloni, Mohamed Elhoseiny, Marcus Rohrbach, and Tinne Tuytelaars.
\newblock Memory aware synapses: Learning what (not) to forget.
\newblock In \emph{ECCV}, pages 139--154, 2018.

\bibitem[Chaudhry et~al.(2018)Chaudhry, Ranzato, Rohrbach, and Elhoseiny]{chaudhry2018efficient}
Arslan Chaudhry, Marc'Aurelio Ranzato, Marcus Rohrbach, and Mohamed Elhoseiny.
\newblock Efficient lifelong learning with a-gem.
\newblock \emph{ICLR}, 2018.

\bibitem[Deng et~al.(2009)Deng, Dong, Socher, Li, Li, and Fei-Fei]{deng2009imagenet}
Jia Deng, Wei Dong, Richard Socher, Li-Jia Li, Kai Li, and Li Fei-Fei.
\newblock Imagenet: A large-scale hierarchical image database.
\newblock In \emph{CVPR}, pages 248--255. Ieee, 2009.

\bibitem[Ding et~al.(2023)Ding, Qin, Yang, Wei, Yang, Su, Hu, Chen, Chan, Chen, et~al.]{ding2023parameter}
Ning Ding, Yujia Qin, Guang Yang, Fuchao Wei, Zonghan Yang, Yusheng Su, Shengding Hu, Yulin Chen, Chi-Min Chan, Weize Chen, et~al.
\newblock Parameter-efficient fine-tuning of large-scale pre-trained language models.
\newblock \emph{Nature Machine Intelligence}, 5\penalty0 (3):\penalty0 220--235, 2023.

\bibitem[Dosovitskiy et~al.(2020)Dosovitskiy, Beyer, Kolesnikov, Weissenborn, Zhai, Unterthiner, Dehghani, Minderer, Heigold, Gelly, et~al.]{dosovitskiy2020image}
Alexey Dosovitskiy, Lucas Beyer, Alexander Kolesnikov, Dirk Weissenborn, Xiaohua Zhai, Thomas Unterthiner, Mostafa Dehghani, Matthias Minderer, Georg Heigold, Sylvain Gelly, et~al.
\newblock An image is worth 16x16 words: Transformers for image recognition at scale.
\newblock \emph{arXiv preprint arXiv:2010.11929}, 2020.

\bibitem[Fernando et~al.(2017)Fernando, Banarse, Blundell, Zwols, Ha, Rusu, Pritzel, and Wierstra]{fernando2017pathnet}
Chrisantha Fernando, Dylan Banarse, Charles Blundell, Yori Zwols, David Ha, Andrei~A Rusu, Alexander Pritzel, and Daan Wierstra.
\newblock Pathnet: Evolution channels gradient descent in super neural networks.
\newblock \emph{arXiv preprint arXiv:1701.08734}, 2017.

\bibitem[Hendrycks et~al.(2021)Hendrycks, Basart, Mu, Kadavath, Wang, Dorundo, Desai, Zhu, Parajuli, Guo, Song, Steinhardt, and Gilmer]{hendrycks2021many}
Dan Hendrycks, Steven Basart, Norman Mu, Saurav Kadavath, Frank Wang, Evan Dorundo, Rahul Desai, Tyler Zhu, Samyak Parajuli, Mike Guo, Dawn Song, Jacob Steinhardt, and Justin Gilmer.
\newblock The many faces of robustness: A critical analysis of out-of-distribution generalization.
\newblock \emph{ICCV}, 2021.

\bibitem[Houlsby et~al.(2019)Houlsby, Giurgiu, Jastrzebski, Morrone, De~Laroussilhe, Gesmundo, Attariyan, and Gelly]{houlsby2019parameter}
Neil Houlsby, Andrei Giurgiu, Stanislaw Jastrzebski, Bruna Morrone, Quentin De~Laroussilhe, Andrea Gesmundo, Mona Attariyan, and Sylvain Gelly.
\newblock Parameter-efficient transfer learning for nlp.
\newblock In \emph{ICML}, pages 2790--2799. PMLR, 2019.

\bibitem[Hu et~al.(2022)Hu, Shen, Wallis, Allen-Zhu, Li, Wang, Wang, Chen, et~al.]{hu2022lora}
Edward~J Hu, Yelong Shen, Phillip Wallis, Zeyuan Allen-Zhu, Yuanzhi Li, Shean Wang, Lu Wang, Weizhu Chen, et~al.
\newblock Lora: Low-rank adaptation of large language models.
\newblock \emph{ICLR}, 1\penalty0 (2):\penalty0 3, 2022.

\bibitem[Kirkpatrick et~al.(2017)Kirkpatrick, Pascanu, Rabinowitz, Veness, Desjardins, Rusu, Milan, Quan, Ramalho, Grabska-Barwinska, et~al.]{kirkpatrick2017overcoming}
James Kirkpatrick, Razvan Pascanu, Neil Rabinowitz, Joel Veness, Guillaume Desjardins, Andrei~A Rusu, Kieran Milan, John Quan, Tiago Ramalho, Agnieszka Grabska-Barwinska, et~al.
\newblock Overcoming catastrophic forgetting in neural networks.
\newblock \emph{Proceedings of the national academy of sciences}, 114\penalty0 (13):\penalty0 3521--3526, 2017.

\bibitem[Krizhevsky et~al.(2009)Krizhevsky, Hinton, et~al.]{krizhevsky2009learning}
Alex Krizhevsky, Geoffrey Hinton, et~al.
\newblock Learning multiple layers of features from tiny images.
\newblock 2009.

\bibitem[Lester et~al.(2021)Lester, Al-Rfou, and Constant]{lester2021power}
Brian Lester, Rami Al-Rfou, and Noah Constant.
\newblock The power of scale for parameter-efficient prompt tuning.
\newblock \emph{arXiv preprint arXiv:2104.08691}, 2021.

\bibitem[Li and Liang(2021)]{li2021prefix}
Xiang~Lisa Li and Percy Liang.
\newblock Prefix-tuning: Optimizing continuous prompts for generation.
\newblock \emph{arXiv preprint arXiv:2101.00190}, 2021.

\bibitem[Li and Hoiem(2017)]{li2017learning}
Zhizhong Li and Derek Hoiem.
\newblock Learning without forgetting.
\newblock \emph{TPAMI}, 40\penalty0 (12):\penalty0 2935--2947, 2017.

\bibitem[Liang and Li(2024)]{liang2024inflora}
Yan-Shuo Liang and Wu-Jun Li.
\newblock Inflora: Interference-free low-rank adaptation for continual learning.
\newblock In \emph{CVPR}, pages 23638--23647, 2024.

\bibitem[Lopez-Paz and Ranzato(2017)]{lopez2017gradient}
David Lopez-Paz and Marc'Aurelio Ranzato.
\newblock Gradient episodic memory for continual learning.
\newblock \emph{Neurips}, 30, 2017.

\bibitem[Loshchilov and Hutter(2017)]{loshchilov2017decoupled}
Ilya Loshchilov and Frank Hutter.
\newblock Decoupled weight decay regularization.
\newblock \emph{arXiv preprint arXiv:1711.05101}, 2017.

\bibitem[Luo et~al.(2012)Luo, Xia, and Zhu]{luo2012incremental}
Xin Luo, Yunni Xia, and Qingsheng Zhu.
\newblock Incremental collaborative filtering recommender based on regularized matrix factorization.
\newblock \emph{Knowledge-Based Systems}, 27:\penalty0 271--280, 2012.

\bibitem[Peng et~al.(2019)Peng, Bai, Xia, Huang, Saenko, and Wang]{peng2019moment}
Xingchao Peng, Qinxun Bai, Xide Xia, Zijun Huang, Kate Saenko, and Bo Wang.
\newblock Moment matching for multi-source domain adaptation.
\newblock In \emph{ICCV}, pages 1406--1415, 2019.

\bibitem[Rebuffi et~al.(2017)Rebuffi, Kolesnikov, Sperl, and Lampert]{rebuffi2017icarl}
Sylvestre-Alvise Rebuffi, Alexander Kolesnikov, Georg Sperl, and Christoph~H Lampert.
\newblock icarl: Incremental classifier and representation learning.
\newblock In \emph{CVPR}, pages 2001--2010, 2017.

\bibitem[Smith et~al.(2023{\natexlab{a}})Smith, Hsu, Zhang, Hua, Kira, Shen, and Jin]{smith2023continual}
James~Seale Smith, Yen-Chang Hsu, Lingyu Zhang, Ting Hua, Zsolt Kira, Yilin Shen, and Hongxia Jin.
\newblock Continual diffusion: Continual customization of text-to-image diffusion with c-lora.
\newblock \emph{arXiv preprint arXiv:2304.06027}, 2023{\natexlab{a}}.

\bibitem[Smith et~al.(2023{\natexlab{b}})Smith, Karlinsky, Gutta, Cascante-Bonilla, Kim, Arbelle, Panda, Feris, and Kira]{smith2023coda}
James~Seale Smith, Leonid Karlinsky, Vyshnavi Gutta, Paola Cascante-Bonilla, Donghyun Kim, Assaf Arbelle, Rameswar Panda, Rogerio Feris, and Zsolt Kira.
\newblock Coda-prompt: Continual decomposed attention-based prompting for rehearsal-free continual learning.
\newblock In \emph{CVPR}, pages 11909--11919, 2023{\natexlab{b}}.

\bibitem[Wah et~al.(2011)Wah, Branson, Welinder, Perona, and Belongie]{WahCUB_200_2011}
C. Wah, S. Branson, P. Welinder, P. Perona, and S. Belongie.
\newblock Technical Report CNS-TR-2011-001, California Institute of Technology, 2011.

\bibitem[Wang et~al.(2024{\natexlab{a}})Wang, Xie, Zhang, Huang, Su, and Zhu]{wang2024hierarchical}
Liyuan Wang, Jingyi Xie, Xingxing Zhang, Mingyi Huang, Hang Su, and Jun Zhu.
\newblock Hierarchical decomposition of prompt-based continual learning: Rethinking obscured sub-optimality.
\newblock \emph{Neurips}, 36, 2024{\natexlab{a}}.

\bibitem[Wang et~al.(2024{\natexlab{b}})Wang, Zhang, Su, and Zhu]{wang2024comprehensive}
Liyuan Wang, Xingxing Zhang, Hang Su, and Jun Zhu.
\newblock A comprehensive survey of continual learning: Theory, method and application.
\newblock \emph{IEEE Transactions on Pattern Analysis and Machine Intelligence}, 2024{\natexlab{b}}.

\bibitem[Wang et~al.(2022{\natexlab{a}})Wang, Liu, Duan, Kong, and Tao]{wang2022continual}
Zhen Wang, Liu Liu, Yiqun Duan, Yajing Kong, and Dacheng Tao.
\newblock Continual learning with lifelong vision transformer.
\newblock In \emph{CVPR}, pages 171--181, 2022{\natexlab{a}}.

\bibitem[Wang et~al.(2022{\natexlab{b}})Wang, Zhang, Ebrahimi, Sun, Zhang, Lee, Ren, Su, Perot, Dy, et~al.]{wang2022dualprompt}
Zifeng Wang, Zizhao Zhang, Sayna Ebrahimi, Ruoxi Sun, Han Zhang, Chen-Yu Lee, Xiaoqi Ren, Guolong Su, Vincent Perot, Jennifer Dy, et~al.
\newblock Dualprompt: Complementary prompting for rehearsal-free continual learning.
\newblock In \emph{ECCV}, pages 631--648. Springer, 2022{\natexlab{b}}.

\bibitem[Wang et~al.(2022{\natexlab{c}})Wang, Zhang, Lee, Zhang, Sun, Ren, Su, Perot, Dy, and Pfister]{wang2022learning}
Zifeng Wang, Zizhao Zhang, Chen-Yu Lee, Han Zhang, Ruoxi Sun, Xiaoqi Ren, Guolong Su, Vincent Perot, Jennifer Dy, and Tomas Pfister.
\newblock Learning to prompt for continual learning.
\newblock In \emph{CVPR}, pages 139--149, 2022{\natexlab{c}}.

\bibitem[Wu et~al.(2025)Wu, Piao, Huang, Wang, Li, Pfister, Meng, Ma, and Wei]{wu2025s}
Yichen Wu, Hongming Piao, Long-Kai Huang, Renzhen Wang, Wanhua Li, Hanspeter Pfister, Deyu Meng, Kede Ma, and Ying Wei.
\newblock Sd-lora: Scalable decoupled low-rank adaptation for class incremental learning.
\newblock \emph{ICLR}, 2025.

\bibitem[Yan et~al.(2021)Yan, Xie, and He]{yan2021dynamically}
Shipeng Yan, Jiangwei Xie, and Xuming He.
\newblock Der: Dynamically expandable representation for class incremental learning.
\newblock In \emph{CVPR}, pages 3014--3023, 2021.

\bibitem[Zenke et~al.(2017)Zenke, Poole, and Ganguli]{zenke2017continual}
Friedemann Zenke, Ben Poole, and Surya Ganguli.
\newblock Continual learning through synaptic intelligence.
\newblock In \emph{ICML}, pages 3987--3995. PMLR, 2017.

\bibitem[Zhou et~al.(2024)Zhou, Sun, Ye, and Zhan]{zhou2024expandable}
Da-Wei Zhou, Hai-Long Sun, Han-Jia Ye, and De-Chuan Zhan.
\newblock Expandable subspace ensemble for pre-trained model-based class-incremental learning.
\newblock In \emph{CVPR}, pages 23554--23564, 2024.

\end{thebibliography}
}


\end{document}